\documentclass[sigconf]{acmart}
\usepackage{listings}
\usepackage{xcolor}
\usepackage{amsfonts}
\usepackage{amsmath}
\usepackage{graphicx}
\usepackage[bb=dsserif]{mathalpha}
\usepackage{bm}
\usepackage{float}

\newcommand{\evosax}{\href{https://github.com/RobertTLange/evosax}{\texttt{evosax}} }

\usepackage{pifont}
\newcommand{\xmark}{\ding{55}}%
\newcommand{\cmark}{\ding{51}}%

\definecolor{codegreen}{rgb}{0,0.6,0}
\definecolor{codegray}{rgb}{0.5,0.5,0.5}
\definecolor{codepurple}{rgb}{0.58,0,0.82}
\definecolor{backcolour}{rgb}{0.95,0.95,0.92}

\usepackage{color, colortbl}
\usepackage[first=0,last=9]{lcg}

\definecolor{LightYellow}{rgb}{0.99, 0.99, 0.59}
\definecolor{LightRed}{rgb}{0.97, 0.51, 0.47}
\definecolor{LightBlue}{rgb}{0.99, 0.59, 0.99}
\definecolor{LightGreen}{rgb}{0.59, 0.99, 0.99}

\definecolor{codegreen}{rgb}{0,0.6,0}
\definecolor{codegray}{rgb}{0.5,0.5,0.5}

\definecolor{backcolour}{RGB}{245,248,250}
\definecolor{emph}{RGB}{166,88,53}
\definecolor{nightblue}{RGB}{9,49,105}
\definecolor{keywords}{RGB}{207,33,46}
\definecolor{lightpurple}{RGB}{130,81,223}

\lstdefinestyle{mystyle}{
    backgroundcolor=\color{backcolour},   
    commentstyle=\color{codegreen},
    keywordstyle=\color{keywords},
    stringstyle=\color{nightblue},
    basicstyle=\ttfamily\footnotesize,
    breakatwhitespace=false,         
    breaklines=true,                 
    captionpos=b,                    
    keepspaces=true,                 
    showspaces=false,                
    showstringspaces=false,
    showtabs=false,                  
    tabsize=2,
    frame=shadowbox,
    emph={AutoTokenizer,AutoModelForSequenceClassification,Explainer},
    emphstyle={\color{emph}},
    emph={[2]from_pretrained,compute_table},
    emphstyle={[2]\color{lightpurple}}
}

\AtBeginDocument{%
  \providecommand\BibTeX{{%
    \normalfont B\kern-0.5em{\scshape i\kern-0.25em b}\kern-0.8em\TeX}}}

\setcopyright{acmcopyright}
\copyrightyear{2022}
\acmYear{2022}
\acmDOI{10.1145/1122445.1122456}

\begin{document}

\title{\evosax: JAX-Based Evolution Strategies}

\author{Robert Tjarko Lange}
\email{robert.t.lange@tu-berlin.de}
\affiliation{%
  \institution{Technical University Berlin}
  \country{Science of Intelligence Cluster of Excellence}
}

\renewcommand{\shortauthors}{Lange}

\begin{abstract}
  The deep learning revolution has greatly been accelerated by the 'hardware lottery': Recent advances in modern hardware accelerators and compilers paved the way for large-scale batch gradient optimization.
  Evolutionary optimization, on the other hand, has mainly relied on CPU-parallelism, e.g. using Dask scheduling and distributed multi-host infrastructure. Here we argue that also modern evolutionary computation can significantly benefit from the massive computational throughput provided by GPUs and TPUs. In order to better harness these resources and to enable the next generation of black-box optimization algorithms, we release \evosax: A JAX-based library of evolution strategies which allows researchers to leverage powerful function transformations such as just-in-time compilation, automatic vectorization and hardware parallelization.\footnote{The library is publicly available at \textbf{\url{https://github.com/RobertTLange/evosax}} under Apache-2.0 license. This report is based on the release version 0.1.0 (December 2022).}
  \evosax implements 30 evolutionary optimization algorithms including finite-difference-based, estimation-of-distribution evolution strategies and various genetic algorithms. Every single algorithm can directly be executed on hardware accelerators and automatically vectorized or parallelized across devices using a single line of code. It is designed in a modular fashion and allows for flexible usage via a simple ask-evaluate-tell API. We thereby hope to facilitate a new wave of scalable evolutionary optimization algorithms.
  
\end{abstract}


\maketitle
\section{Introduction}

The exponential growth of general-purpose compute resources (Moore's law) has slowed down. Researchers may no longer expect an exponential increase in general compute capabilities. Instead, more and more application-specific devices (ASICs) are being developed in order to meet the demands of both practitioners and researchers. These tailored hardware advances have arguably enabled and accelerated the recent success of deep learning and large-scale gradient descent-based (GD) optimization \citep{hooker2021hardware}. 

But how can evolution strategies (ES) fully leverage modern day ASICs? ES perform black-box optimization (BBO) via sequential evaluation of batches of candidate solutions. In order to estimate the fitness of each population member, one often times require the execution of several stochastic function evaluations (e.g. agent policy rollouts or batched loss computations).
One obvious application of recent hardware advances therefore is the distributed evaluation of population members \citep{freeman2021brax, gymnax2022github}. But there exists another use case: The acceleration and parallelization of the execution of the EO algorithm itself -- directly on the accelerator. Here, we argue that JAX \citep{jax2018github} provides a general tool for this purpose, which opens up many new research directions for evolutionary optimization. We introduce \evosax, a JAX-based library of ES, which implements implements 30 evolutionary optimization (EO) algorithms and comes with the following core features:

\begin{enumerate}
    \item \textbf{Flexibility}: It implements JAX-transformable evolution strategies and covers many classic ES (e.g. CMA-ES \citep{hansen2001completely}) and modern neuroevolution ES (e.g. ARS \citep{mania2018simple}, OpenAI-ES \citep{salimans2017evolution}).
    \item \textbf{Acceleration}: All ES sampling \& update steps can be just-in-time compiled and executed on different hardware devices.
    \item \textbf{Parallelization}: A single line of code allows the user to execute multiple ES runs in parallel for different random seeds or hyperparameter settings. 
    \item \textbf{Modularity}: We provide utilities for automated parameter reshaping, fitness shaping, network definition, accelerated problem rollout \& restart wrappers. All of these can be combined in a modular fashion and allow for design flexibility.
\end{enumerate}

\begin{figure}[H]
\centering
\includegraphics[width=0.475\textwidth]{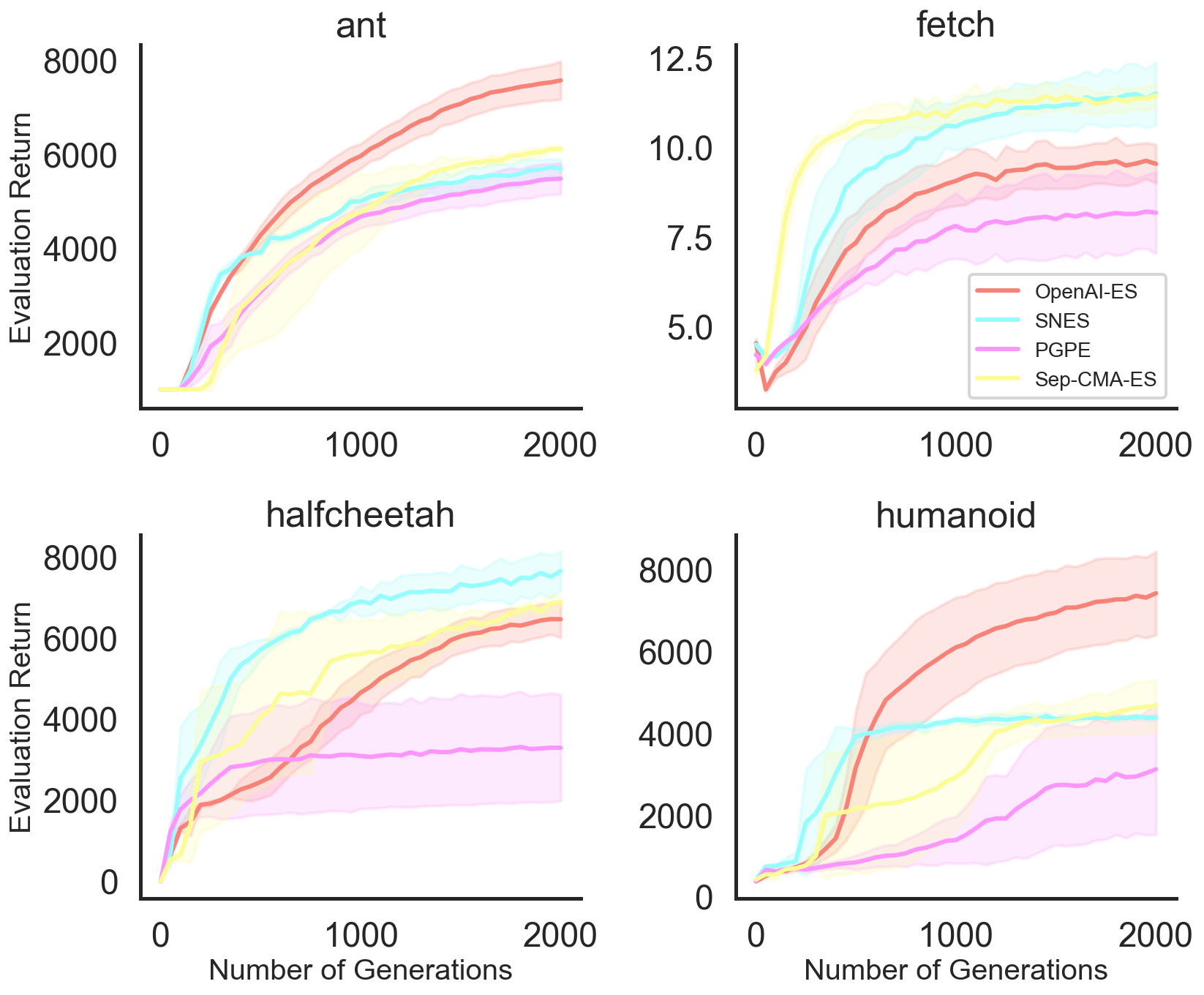}%
\caption{Comparison of different evolution strategies (OpenAI-ES, PGPE, SNES and Sep-CMA-ES) implemented by \evosax on four continuous control tasks using the Brax library \citep{freeman2021brax}. The simulations use a population size of 256 and the results are averaged over 3 independent runs and we report one standard deviation bars.}
\label{fig1:curves}
\end{figure}

\begin{figure}[h]
\centering
\includegraphics[width=0.475\textwidth]{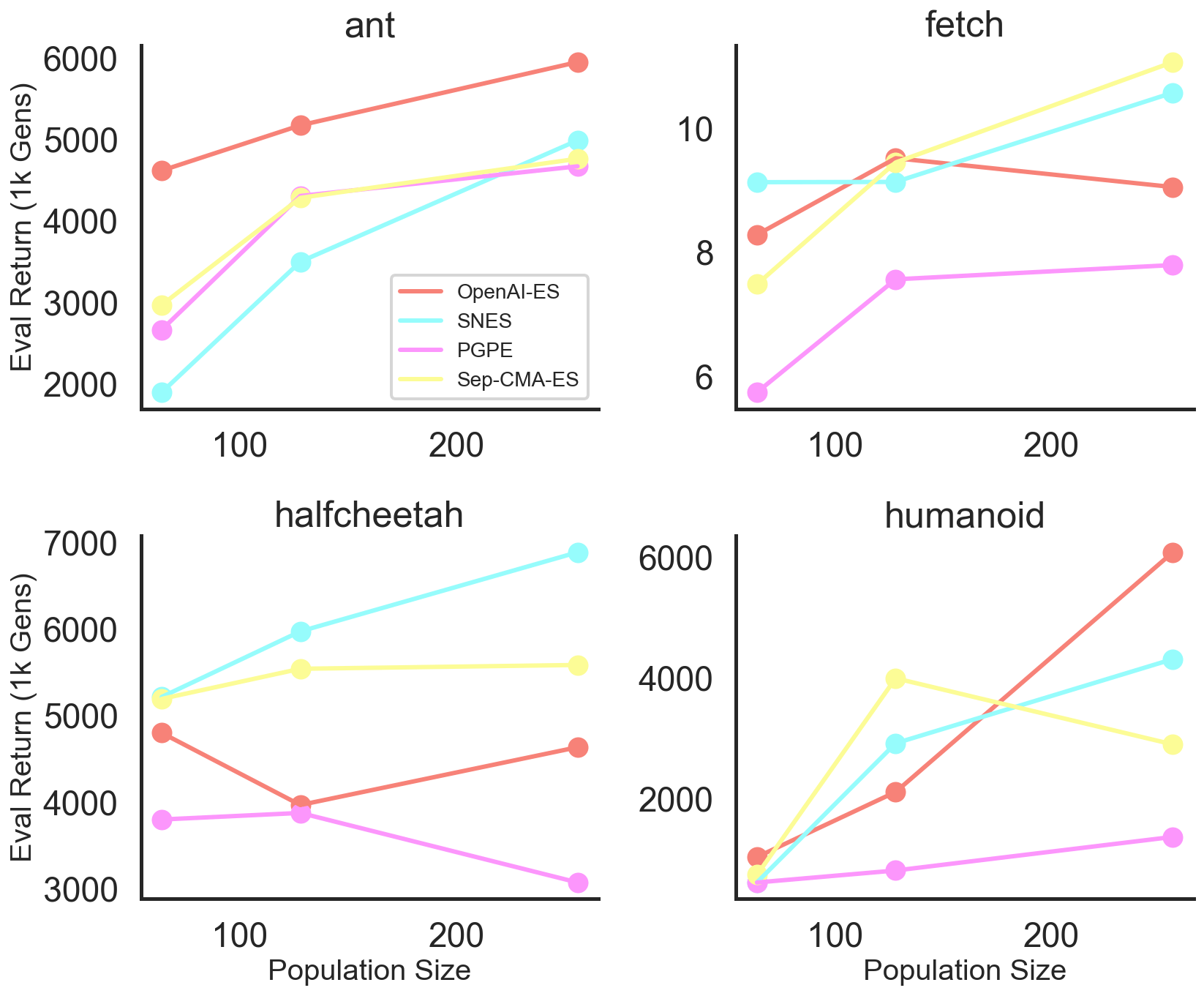}%
\caption{Comparison of different ES (OpenAI-ES, PGPE, SNES and Sep-CMA-ES) across population sizes on four continuous control tasks. The results are averaged over 3 independent runs.}
\label{fig2:populations}
\end{figure}

\begin{figure*}[h]
\centering
\includegraphics[width=0.95\textwidth]{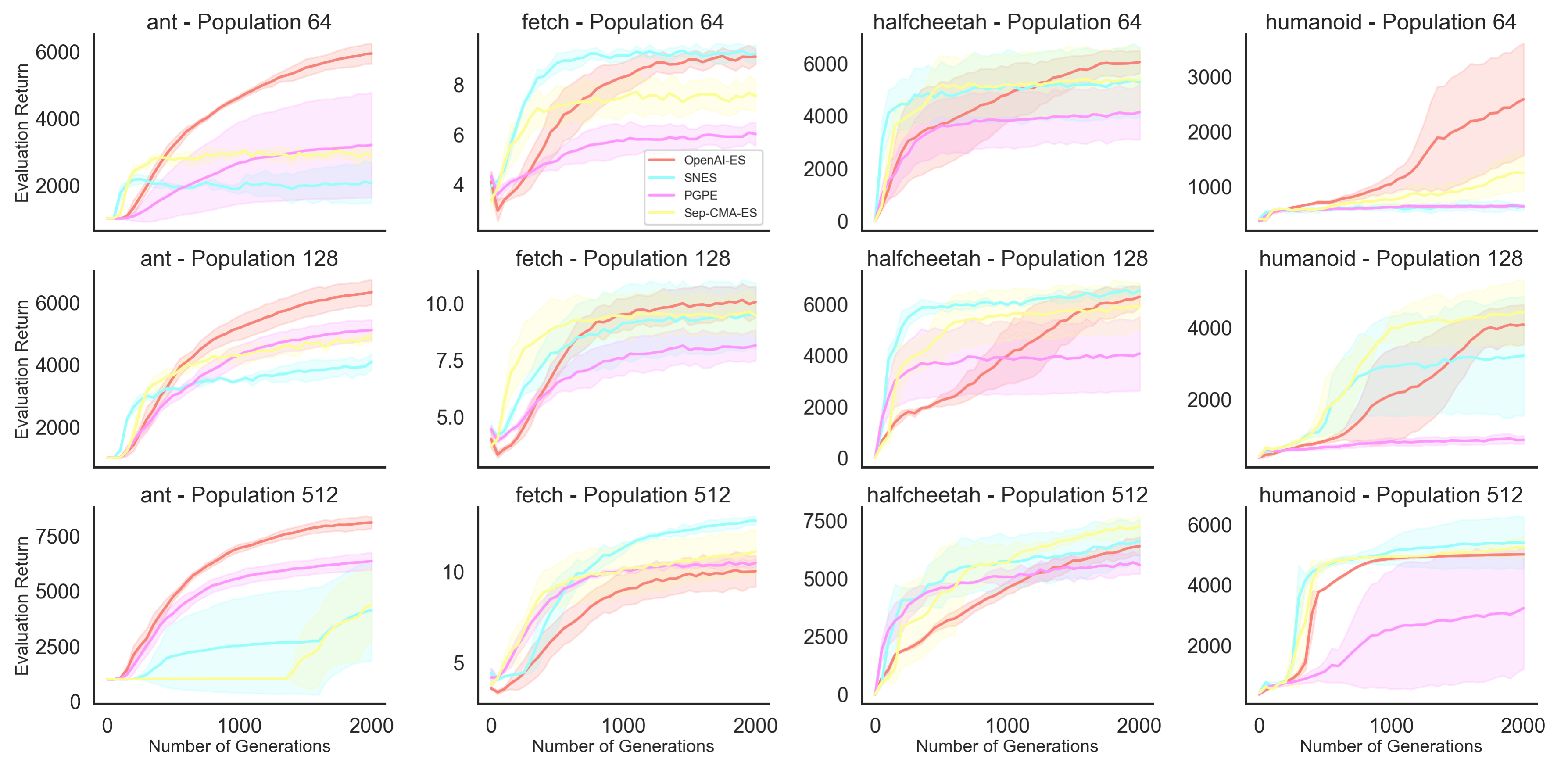}%
\caption{Comparison of different ES (OpenAI-ES, PGPE, SNES and Sep-CMA-ES) across population sizes on four continuous control tasks. The results are averaged over 3 independent runs and we report one standard deviation bars.}
\label{fig3:details}
\end{figure*}

Finally, we provide the results for experiments conducted on four neuroevolution tasks comparing representative ES (see Figures \ref{fig1:curves}, \ref{fig2:populations}, \ref{fig3:details}). We hope that these contributions can enable new research directions including evolutionary meta-learning, the discovery of neural network-parametrized ES, sub-population management and strategy ensembling. 

\newpage
\section{Background}

\subsection{BBO via Evolution Strategies}

\textbf{Continuous Black-Box Optimization.}
Throughout this manuscript, we are interested in efficient black-box optimization: Given a function $f(x): \mathbb{R}^D \to \mathbb{R}$ with unknown functional form, i.e. we cannot compute its derivative, we seek to find its global optimum:
\vspace{-0.05cm}
$$\min_\mathbf{x} f(\mathbf{x}), \text{s.t.} \ \mathbf{x}_d \in [l_d, u_d] \subset [-\infty, \infty], \forall d=1,...,D.$$
\vspace{-0.35cm}
\\
\textbf{Evolution Strategies.} ES maintain a search distribution and aim to maximize the probability of sampling high fitness candidate solutions. Most common ES assume the search distribution to be a multivariate Gaussian distribution parameterized by sufficient statistics consisting of a mean $\mathbf{m} \in \mathbb{R}^D$ vector and covariance matrix $ \boldsymbol{\Sigma} \in \mathbb{R}^{D \times D}$, where $D$ denotes the number of search space dimensions. At each generation one samples a population of candidate solutions $\mathbf{x}_j \in \mathbb{R}^D$ for all $j = 1, ... , N$ population members and evaluates their fitness $f(\mathbf{x}_j)$. Afterwards, the search distribution is updated to increase the likelihood of well-performing solutions. Roughly speaking, ES differ in the assumed covariance structure and the computation of an 'evolutionary gradient'. The ES provided by \evosax can be grouped into three classes:

\begin{itemize}
    \item \textbf{[FDG]} Finite-difference (FD) gradient ES: One subclass uses random perturbations to calculate an estimate of the FD fitness gradient \citep{salimans2017evolution}:
\vspace{-0.1cm}
	$$\nabla_\mathbf{m} \mathbb{E}_{\epsilon \sim \mathcal{N}(0, I)} f(\mathbf{m} + \sigma \epsilon) = \frac{1}{\sigma} \mathbb{E}_{\epsilon \sim \mathcal{N}(0, I)} [f(\mathbf{m} + \sigma \epsilon) \epsilon].$$
	This estimate is then used along standard GD-based optimizers (e.g. Adam) to update the search mean $\mathbf{m}$. FDG-ES differ in their use of fitness shaping, elite selection and covariance structure (isotropic \citep{salimans2017evolution, mania2018simple} or diagonal \citep{sehnke2010parameter}) and often make use of antithetic sampling for variance reduction.
    \item \textbf{[PFD]} Natural evolution strategies and pre-conditioning: PFD methods extend the FD approach to ES by incorporating uncertainty into the estimation of the fitness gradient. More specifically, they pre-condition the gradient updates using an estimate of the inverse Fisher matrix \citep{wierstra2014natural} or rely on gradient subspace-constrained candidate sampling \citep{choromanski2019complexity}. 
    \item \textbf{[EOD]} Estimation of distribution: Algorithms such as CMA-ES \citep{hansen2001completely} and its diagonal version, Sep-CMA-ES \citep{ros2008simple}, rely on weighted recombination-based mean updates and iterative covariance matrix estimation. They rely on so-called evolution paths that estimate momentum-like statistics to guide the structure of the covariance.\\
\end{itemize}
\vspace{-0.45cm}
\textbf{\texttt{ask} - \texttt{evaluate} - \texttt{tell} API.} \evosax adopts a common API implemented by most ES software packages. More specifically, it follows a three step paradigm, which is iterated for each generation:

\begin{enumerate}
    \item \textbf{ask}: Given the sufficient statistics of the search distribution, one samples a set of candidate solutions: E.g. $\mathbf{x}_j \in \mathbb{R}^D \sim \mathcal{N}(\mathbf{m}, \boldsymbol{\Sigma})$ for all $j = 1, ... , N$ (\texttt{strategy.ask(...)}).
    \item \textbf{evaluate}: Afterwards, all $N$ candidates are evaluated on the black-box task of interest, potentially multiple times to obtain their fitness estimates, $\hat{f}_j$ for all $j = 1, ... , N$.
    \item \textbf{tell}: The collected information is then used to update the search distribution ($\mathbf{m}', \mathbf{\Sigma}' \leftarrow \texttt{ES-UPDATE}(\mathbf{m}, \mathbf{\Sigma}, \{\mathbf{x}_j\}, \{\hat{f}_j\}$) used in the next generation (\texttt{strategy.tell(...)}).
\end{enumerate}

Listing \ref{api-evosax} summarizes how \evosax implements this procedure:

\begin{lstlisting}[language=Python, caption=ask-evaluate-tell API for \evosax (e.g. CMA-ES)., label=api-evosax]
import jax
from evosax import CMA_ES

# Instantiate the search strategy & hyperparameters
rng = jax.random.PRNGKey(0)
strategy = CMA_ES(popsize=20, num_dims=2)
es_params = strategy.default_params
es_state = strategy.initialize(rng, es_params)

# Run ask-eval-tell loop - NOTE: By default minimization!
for t in range(num_generations):
  rng, rng_gen, rng_eval = jax.random.split(rng, 3)
  x, es_state = strategy.ask(rng_gen, es_state, es_params)
  fitness = ...  # Your evaluation fct for x 
  es_state = strategy.tell(x, fitness, es_state, es_params)

# Get search mean, best overall candidate & its fitness
es_state.mean, es_state.best_member, es_state.best_fitness
\end{lstlisting}

\subsection{Evolution Strategies in JAX}

JAX \citep{jax2018github} provides a set of composable function transformations, which easily translate across different hardware devices (CPU/GPU/TPU). It originally became popular as an alternative automatic differentiation toolbox powering distributed deep learning applications. But what can JAX implementations of ES enable for the evolutionary optimization community?\\
\textbf{Just-in-time compiled ES sampling \& updates via \texttt{jax.jit}.}
While often times large chunks of the ES runtime are spend on the evaluation of candidate solutions, sampling of high-dimensional vectors can become a burden. Most ES rely on a form of reparametrization of isotropic Gaussian noise, which can require the Cholesky decomposition of a covariance matrix. This operation can naturally be sped up on accelerators. Furthermore, JAX provides a simple jit-compilation transformation, which further improves running time. Thereby, \evosax-based evolution strategies can naturally scale to larger search dimensions with minimal engineering efforts. Listing \ref{jit-evosax} shows how to compile a full ES optimization loop:

\begin{lstlisting}[language=Python, caption=jit-compiled ES loop execution in \evosax., label=jit-evosax]
@partial(jax.jit, static_argnums=(1,))
def run_es_loop(rng, num_generations):
  """Run evolution ask-eval-tell loop."""
  es_params = strategy.default_params
  es_state = strategy.initialize(rng, es_params)

  def es_step(state_input, tmp):
    """Helper es step to lax.scan through."""
    rng, es_state = state_input
    rng, rng_iter = jax.random.split(rng)
    x, es_state = strategy.ask(rng_iter, es_state, es_params)
    fitness = ...  # Your evaluation fct for x
    es_state = strategy.tell(y, fitness, es_state, es_params)
    return [rng, es_state], fitness[jnp.argmin(fitness)]

  _, scan_out = jax.lax.scan(es_step,
                             [rng, es_state],
                             [jnp.zeros(num_generations)])
    return jnp.min(scan_out)
\end{lstlisting}
\textbf{Automatic parallel ES execution via \texttt{jax.vmap}/\texttt{jax.pmap}.}
Next to jit-compilation and automatic differentiation utilities, JAX additionally provides function transformations, which allow for parallel execution of multiple algorithm instances. This enables the simultaneous evaluation of multiple different ES rollouts on a single device (\texttt{jax.vmap}) and on multiple devices (\texttt{jax.pmap}). A sample of potential applications include the parallel execution of multiple independent runs of the same ES (e.g. by batching over different random number keys), ES characterized by different hyperparameters of the same shape and the design of ES, which operate over sub-populations. Note that these parallel execution can again be composed. 
Listing \ref{map-evosax} shows an example of full ES optimization loop:

\begin{lstlisting}[language=Python, caption=Parallel ES execution in \evosax using \texttt{vmap}/\texttt{pmap}., label=map-evosax]
# Automatic vectorization across random seeds
seed_rng = jax.random.split(rng, num_seeds)
vec_min = jax.vmap(run_es_loop, in_axes=(0, None, None))(seed_rng, num_gens)

# Device parallel execution across random seeds
map_min = jax.pmap(run_es_loop, in_axes=(0, None, None))(seed_rng, num_gens)
\end{lstlisting}
\vspace{-0.4cm}
\subsection{Related Work \& Open Source Projects}

\textbf{JAX-based neuroevolution in \href{https://github.com/google/evojax}{\texttt{EvoJAX}}}. \citet{tang2022evojax} previously introduced a library for neuroevolution powered by JAX. \texttt{EvoJAX} focuses on parallel fitness rollouts, which achieves drastic speed-ups in fitness estimation for neural network-based tasks. They provide utilities for powering ES pipelines and accelerated tasks. Both packages are complementary in that \evosax provides the ES backend for a substantial amount of ES used by \texttt{EvoJAX} at the time of writing. Furthermore, \evosax comes with a simple to use wrapper that allows users to combine \evosax ES with EvoJAX rollouts.\\
\textbf{JAX-based Quality-Diversity algorithms in \href{https://github.com/adaptive-intelligent-robotics/QDax}{\texttt{qdax}}}. \citet{lim2022accelerated} further provide Quality-Diversity (QD) algorithms powered by JAX. QD consistute a set of evolutionary algorithms, which jointly optimize fitness and diversity of solution vectors, measured by a set of behavioral descriptors. Again, JAX provides improvements in fitness/descriptor acquisition and can speed up candidate emititation run times. \evosax can in principle be combined with the mutation and candidate proposal mechanisms of \texttt{qdax}.\\
\textbf{JAX-based Reinforcement Learning environments}. Finally, \evosax can easily be used with a set of recently proposed JAX-based control tasks. These include continuous control tasks implemented by \href{https://github.com/google/brax/}{\texttt{Brax}} \citep{freeman2021brax}, classic gym control tasks implemented by \href{https://github.com/RobertTLange/gymnax}{\texttt{gymnax}} \citep{gymnax2022github} and industry-specific applications \citep[\href{https://github.com/instadeepai/jumanji}{\texttt{Jumanji}},][]{jumanji2022github}.

\section{\evosax: ES, Problems \& Utilities}

\subsection{Accelerated Evolution Strategies}
At its core \evosax implements a diverse set of ES, which can leverage the previously reviewed function transformations provided by JAX. We further include gradient-based optimizers and their ES adaptations (e.g. the ClipUp optimizer \citep{toklu2020clipup}). Additionally, we provide a set of representative genetic algorithms \textbf{[GA]} and miscellaneous evolutionary optimization algorithms \textbf{[MEO]} in order to facilitate benchmark comparisons across EO algorithm classes. In total the package provides a set of 30 evolutionary optimization algorithms summarized in Table \ref{strategy-table}:
\begin{table}[h]
  \caption{Evolutionary optimizers implemented by \evosax}
  \label{strategy-table}
  \centering
  \begin{tabular}{l|ll|cc}
    \toprule
    \multicolumn{3}{c}{Evolution Strategy}  &  \multicolumn{2}{c}{\texttt{ask}/\texttt{tell}}                 \\
    \cmidrule(r){1-4}\cmidrule(r){4-5}
    Class & Name     & Reference     & \texttt{jit} & \texttt{map} \\
    \midrule
    \rowcolor{LightRed}
    FDG & OpenAI-ES & \citet{salimans2017evolution}  & \cmark & \cmark     \\
    \rowcolor{LightBlue}
    FDG & PGPE & \citet{sehnke2010parameter}  & \cmark & \cmark     \\
    FDG & ARS & \citet{mania2018simple}  & \cmark & \cmark     \\
    FDG & Persistent ES & \citet{vicol2021unbiased}  & \cmark & \cmark     \\
    FDG & ESMC & \citet{merchant2021learn2hop} & \cmark & \cmark \\
    \hline
    PFD & xNES & \citet{wierstra2014natural}  & \cmark & \cmark     \\
    \rowcolor{LightGreen}
    PFD & SNES & \citet{schaul2011high}  & \cmark & \cmark     \\
    PFD & CR-FM-NES & \citet{nomura2022fast} & \cmark & \cmark     \\
    PFD & Guided ES & \citet{maheswaranathan2019guided} & \cmark & \cmark \\
    PFD & ASEBO & \citet{choromanski2019complexity} & \cmark & \cmark \\
    \hline
    EOD & CMA-ES & \citet{hansen2001completely}  & \cmark & \cmark     \\
    \rowcolor{LightYellow}
    EOD & Sep-CMA-ES & \citet{ros2008simple}  & \cmark & \cmark     \\
    EOD & IPOP-CMA & \citet{auger2005restart}  & (\cmark) & \xmark     \\
    EOD & BIPOP-CMA & \citet{hansen2009benchmarking}  & (\cmark) & \xmark     \\
    EOD & RmES & \citet{li2017simple}  & \cmark & \cmark     \\
    EOD & MA-ES & \citet{beyer2017simplify}  & \cmark & \cmark     \\
    EOD & LM-MA-ES & \citet{loshchilov2017limited}  & \cmark & \cmark     \\
    EOD & (i)AMaLGaM & \citet{bosman2013benchmarking}  & \cmark & \cmark     \\
    EOD & Gaussian ES & \citet{rechenberg1978evolutionsstrategien}  & \cmark & \cmark     \\
    EOD & Discovered ES & \citet{lange2022discovering}  & \cmark & \cmark     \\
    \hline
    GA & Gaussian GA & \citet{such2017deep}  & \cmark & \cmark     \\
    GA & SAMR-GA & \citet{clune2008natural}  & \cmark & \cmark     \\
    GA & GESMR-GA & \citet{kumar2022effective} & \cmark & \cmark     \\
    GA & MR-1/5-GA & \citet{rechenberg1978evolutionsstrategien}  & \cmark & \cmark     \\
    \hline
    MEO & PSO & \citet{kennedy1995particle}  & \cmark & \cmark     \\
    MEO & DE & \citet{storn1997differential}  & \cmark & \cmark     \\
    MEO & GLD & \citet{golovin2019gradientless}  & \cmark & \cmark     \\
    MEO & SimAnneal & \citet{rere2015simulated} & \cmark & \cmark     \\
    MEO & PBT & \citet{jaderberg2017population}  & \cmark & \cmark     \\
    \bottomrule
  \end{tabular}
\end{table}

\subsection{Tasks \& Utilities: Benchmarks, Wrappers, Parameter Reshaping \& Fitness Shaping}

Next to the core algorithms, \evosax comes with a set benchmarking tasks and common utilities. These include the standard BBO benchmark functions \citep[BBOB,][]{hansen2010real}, rollout wrappers for \texttt{gymnax} tasks, common computer vision classification tasks (MNIST, etc.) and sequential prediction tasks. Furthermore, we provide simple neural network architectures (MLP, CNN, LSTM), ES logging and visualization tools.
ES updates commonly operate on flat parameter vectors, while neural networks rely on layer-wise parameter data structures. In order to facilitate the ES workflow, we therefore also provide a tool for effortless parameter reshaping. \texttt{evosax} also comes with a set of different fitness shaping transformations (z-scoring, centered rank computation and weight decay regularization) as well as restart and batch rollout wrappers. 
Finally, ES tend to struggle with high-dimensional optimization problems and can leverage so-called indirect encoding methods to optimize solutions in a lower dimensional subspace. \texttt{evosax} provides different solution vector encodings such as fixed random matrix encodings and hypernetworks \citep{ha2016hypernetworks} for MLP networks. 


\section{Discussion}

\textbf{On the promises of ES.}
While GD-based optimization remains the de facto standard for neural network training, there remain several promising applications for ES in the context of machine learning, e.g. non-differentiable objectives, network operations and architecture search. Furthermore, ES can optimize dynamical systems, which require differentiation through long unrolled computation graphs \citep{metz2022practical, metz2021gradients}. This already enables several instances of evolutionary meta-learning \citep{lange2022discovering, lu2022discovered, lu2022adversarial}. Finally, EO can naturally be applied to delayed reward tasks and long-horizon credit assignment.\\
\textbf{Summary.} We introduced \evosax -- a JAX-based library of EO algorithms, which leverage powerful function transformations to speed up and parallelize execution on modern hardware accelerators. The implementations and accompanying utilities are modular and allow for easy extension and refinement. 
We hope that these contributions can enable a new wave of scalable meta-learned BBO algorithms (e.g. learned ES \citep{lange2022discovering}), which enable applications which are out of reach for standard gradient-based optimization.\\
\textbf{Limitations.}
Current ES struggle with scaling to large search spaces ($D>100$k parameters). This is due to several key challenges including the memory constraints of modern accelerators, the curse of dimensionality affecting effective evolutionary gradient estimation and effective candidate sampling. \\
\textbf{Future Work.} We aim to extend \evosax to allow the user to directly sample sub-populations on different devices and to compute batch evolutionary gradients, which can be aggregated via map reduce-style operations (e.g. \texttt{jax.lax.pmean}). Thereby, the memory requirements of the entire population can be divided across devices. Finally, we will enhance the indirect encodings to include HyperNEAT and wavelet-based encodings \citep{stanley2009hypercube, van2016wavelet}.


\newpage
\bibliographystyle{ACM-Reference-Format}
\renewcommand*{\bibfont}{\tiny}
\bibliography{main}

\section*{Acknowledgments}

We thank Yujin Tang, Yingtao Tian and Chris Lu for several constructive discussions. This work is funded by the Deutsche
Forschungsgemeinschaft (DFG, German Research Foundation) under Germany’s Excellence Strategy - EXC 2002/1 “Science of Intelligence” - project number 390523135. RTL was additionally supported by a Google Cloud Research Credit Grant.

\appendix
\vspace{-0.15cm}
\section{Software \& Compute Requirements}

Simulations were conducted on a high-performance cluster using three independent runs (random seeds). Depending on the setting, a single run lasts between 30 minutes and 4 hours of training time on a single V100S. Experiments were organized using the \texttt{MLE-Infrastructure} \citep[MIT license]{mle_infrastructure2021github} training management system. 

\vspace{-0.15cm}
\section{Brax-Task Hyperparameters}

\begin{table}[H]
\centering
\small
\begin{tabular}{|r|l||r|l|}
\hline \hline
Parameter & Value & Parameter & Value \\ 
\hline
Population Size                & 256   & \# Generations        & 2000      \\
Episode Env Steps                & 1000   & \# Train MC Rollouts        & 16      \\
Evaluate Every Gens                & 25   & \# Test MC Rollouts        & 128      \\
Network Type                & MLP   & \# Hidden Units        & 32      \\
Hidden Activation           & Tanh   & \# Hidden Layers        & 4      \\
Output Activation           & Tanh   & \# Total Params        & 6248      \\
\hline \hline
\end{tabular}
\caption{Shared hyperparameters for Brax \citep{freeman2021brax} control tasks.}
\end{table}

\vspace{-1.05cm}

\begin{table}[H]
\centering
\small
\begin{tabular}{|r|l|r|l|r|}
\hline \hline
Parameter & Ant & Fetch & HalfCheetah & Humanoid \\ 
\hline
Sigma init                 & 0.05   &   0.125      & 0.05 & 0.1      \\
Elite ratio                & 0.4   &     0.2    & 0.5 & 0.2      \\
\hline \hline
\end{tabular}
\caption{Hyperparameters for Sep-CMA-ES \citep{ros2008simple}.}
\end{table}

\vspace{-1.15cm}

\begin{table}[H]
\centering
\small
\begin{tabular}{|r|l|r|l|r|}
\hline \hline
Parameter & Ant & Fetch & HalfCheetah & Humanoid \\ 
\hline
Lrate init                & 0.01   &    0.02     & 0.01  &    0.02\\
-- decay                & 0.999   &    0.999     & 0.999    &  0.999\\
-- final                & 0.001   &  0.001       & 0.001      & 0.001\\
Sigma init                & 0.05   &   0.05      & 0.075   &   0.1\\
-- decay                & 0.999   &   0.999      & 0.999    & 0.999 \\
-- final                & 0.01   &    0.01     & 0.01    &  0.01\\
Optimizer                & Adam   &    Adam     & Adam & Adam    \\
Fitness shaping                & Cent. rank   &    Cent. rank     & Cent. rank   & Cent. rank  \\
\hline \hline
\end{tabular}
\caption{Hyperparameters for OpenAI-ES \citep{salimans2017evolution}.}
\end{table}

\vspace{-1.05cm}

\begin{table}[H]
\centering
\small
\begin{tabular}{|r|l|r|l|r|}
\hline \hline
Parameter & Ant & Fetch & HalfCheetah & Humanoid \\ 
\hline
Lrate init                & 0.01   &    0.02     & 0.02    & 0.02 \\
-- decay                & 0.999   &     0.999    & 0.999     & 0.999\\
-- final                & 0.001   &   0.001      & 0.001    &  0.001\\
Sigma init                & 0.025   &   0.05      & 0.025     & 0.05\\
-- decay                & 0.999   &   0.999      & 0.999     & 0.999 \\
-- final                & 0.01   &     0.01    & 0.01     & 0.01\\
-- learning rate & 0.2 & 0.2& 0.2& 0.2\\
-- max. change & 0.2 & 0.2 & 0.2 & 0.2\\
Optimizer                & Adam   &    Adam     & Adam & Adam    \\
Fitness shaping                & Cent. rank   &    Cent. rank     & Cent. rank   & Cent. rank  \\
\hline \hline
\end{tabular}
\caption{Hyperparameters for PGPE \citep{sehnke2010parameter}.}
\end{table}

\vspace{-1.05cm}

\begin{table}[H]
\centering
\small
\begin{tabular}{|r|l|r|l|r|}
\hline \hline
Parameter & Ant & Fetch & HalfCheetah & Humanoid \\ 
\hline
Sigma init                & 0.05   &     0.075    & 0.05  & 0.075

aa\\
$\beta$ Temperature                & 12   &     12    &  16 & 32      \\
\hline \hline
\end{tabular}
\caption{Hyperparameters for SNES \citep{wierstra2014natural}.}
\end{table}

\vspace{-0.5cm}
SNES uses the following utility/fitness shaping transformation:
\vspace{-0.1cm}
\begin{equation*}
    \boldsymbol{w}_{t, j} = \text{softmax}\left(\beta \times \left(\text{rank}(j)/N - 0.5 \right)\right), \ \forall j=1,..., N.
\end{equation*}

\end{document}